\documentclass[10pt,twocolumn,letterpaper]{article}
\pdfoutput=1  % to ensure that the .tex file is built using PDFLaTeX instead of LaTeX:
\usepackage{wacv}
\usepackage{times}
\usepackage{epsfig}
\usepackage{graphicx}
\usepackage{amsmath}
\usepackage{amssymb}

\graphicspath{{figures/}}
\usepackage{subcaption} % subfigure
\usepackage{caption}
\usepackage{eucal}  % math symbo \mathcal \CMcal
\usepackage{bm}   % use \hl
\DeclareMathOperator*{\argmax}{arg\,max}
\usepackage{soul,color}   % use \hls
\usepackage{multirow}  % for table
\usepackage{booktabs}  % for table rules
\usepackage{array}

\usepackage[title]{appendix}
\usepackage[font=small,skip=0pt]{caption}  % make gap smaller
\usepackage[title]{appendix}

\def\wacvPaperID{544} % Enter the WACV Paper ID here

\wacvfinalcopy % *** Uncomment this line for the final submission

\ifwacvfinal
\def\assignedStartPage{1} % *** Enter the assigned starting page number (instead of 9876)
\fi

\ifwacvfinal
\usepackage[pagebackref=true,breaklinks=true,colorlinks,bookmarks=false]{hyperref}
\else
\usepackage[pagebackref=true,breaklinks=true,colorlinks,bookmarks=false]{hyperref}
\fi

\ifwacvfinal
\else
\fi

\begin{document}

%%%%%%%%% TITLE
\title{An Autonomous Approach to Measure Social Distances and Hygienic Practices during COVID-19 Pandemic in Public Open Spaces}

\author{Peng Sun\\
University of Central Florida\\
{\tt\small peng.sun@ucf.edu}

\and
Gabriel Draughon\\
University of Michigan\\
{\tt\small draughon@umich.edu}

\and
Jerome Lynch\\
University of Michigan\\
{\tt\small jerlynch@umich.edu}
}

\maketitle
%\thispagestyle{empty}

%%%%%%%%% ABSTRACT
\begin{abstract}
  Coronavirus has been spreading around the world since the end of 2019. Due to its high risk of danger causing severe acute respiratory syndrome that could be lethal, human behavior and activities in a crowded environment (e.g. public open spaces) need to be studied. Hence, there is a great need to ensure the safety of public use of park facilities in public open spaces where people's activity and density might be higher due to the needs after stay-at-home executive orders. This work provides a scalable sensing approach to detect physical activities within public open spaces and monitor adherence to social distancing guidelines suggested by the US Centers for Disease Control and Prevention (CDC). A deep learning-based computer vision sensing framework is designed to investigate the careful and proper utilization of parks and their facilities with hard surfaces (e.g. benches, fence poles, and trash cans) using video feeds from a pre-installed surveillance camera network. The sensing framework consists of CNN-based object detector, multi-target tracker, mapping module, and a group reasoning module.  Recognition modules of hard-surface contact and facial clothing/ mask wearing are also incorporated to further analyze the ratio of park users who perform hygienic practices in public spaces as suggested by CDC. The experiments are carried out at several key locations at the Detroit Riverfront Parks between March 2020 and May 2020 during the outbreak of the COVID-19 pandemic in the state of Michigan and Wayne County. The sensing framework is validated by comparing automatic sensing results with manually labeled ground truth results.  The proposed approach significantly improves the efficiency of providing spatial and temporal statistics of users in public open spaces by creating straightforward data visualizations for federal and state agencies. The results can further provide on-time triggering information for an alarming or actuator system which can later be added to intervene inappropriate behavior during this pandemic.
\end{abstract}

%%%%%%%%% BODY TEXT
\section{Introduction}

\begin{figure} [tp]
\vspace{-1.0em}   % make gap between caption and
    \centering
    \begin{subfigure}[b]{0.23\textwidth}
    \centering
    \includegraphics[width=\textwidth]{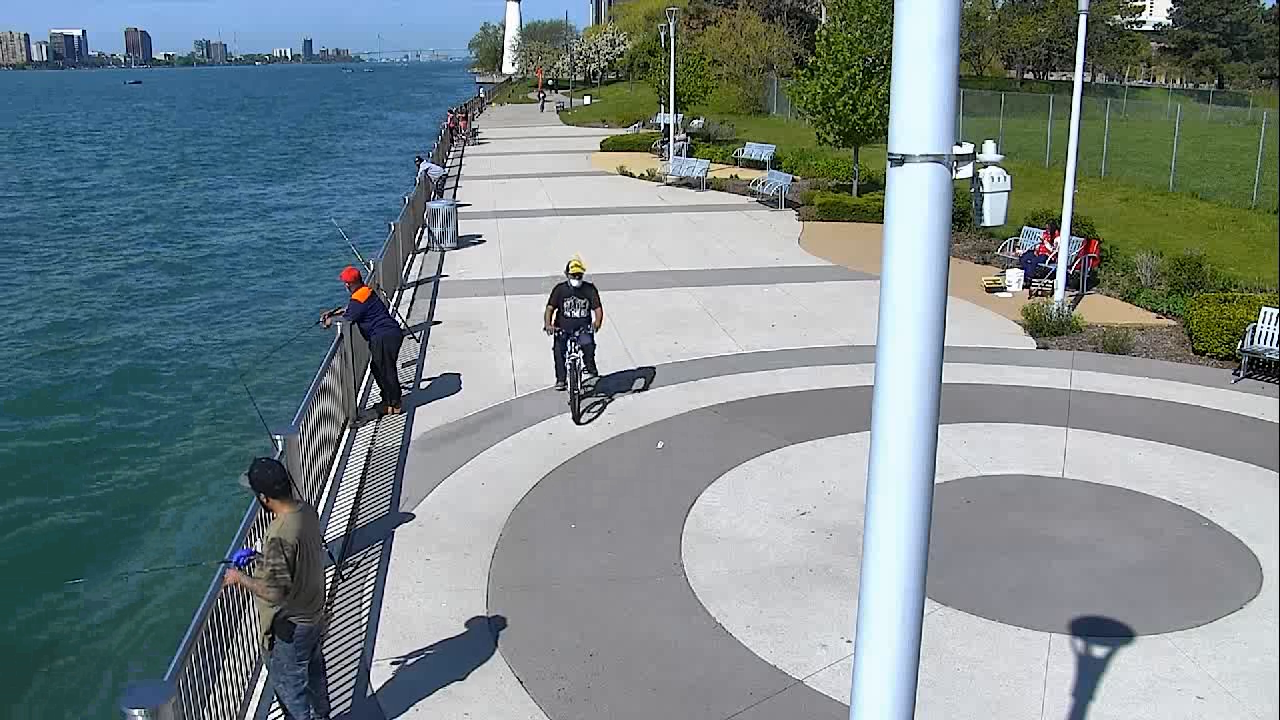}
    \vspace{-1.5em}   % make gap between caption and text smaller
    \caption{Bicycling and fishing}    
    \label{fig:dc_orig_sparse}
    \end{subfigure}
    \,
    \begin{subfigure}[b]{0.23\textwidth}  
    \centering 
    \includegraphics[width=\textwidth]{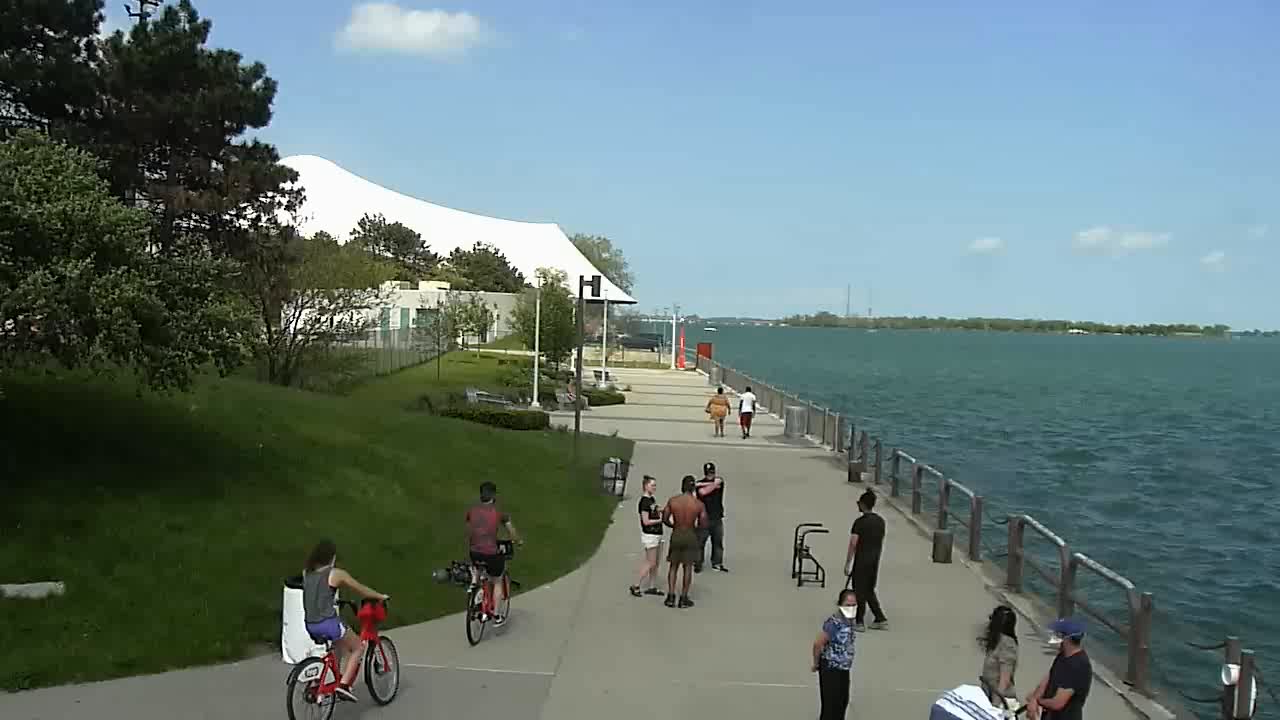}
    \vspace{-1.5em}   % make gap between caption and text smaller
    \caption{Talking between friends}    
    \label{fig:dc_orig_dense}
    \end{subfigure}
    \\
    \begin{subfigure}[b]{0.23\textwidth}
    \centering
    \includegraphics[width=\textwidth]{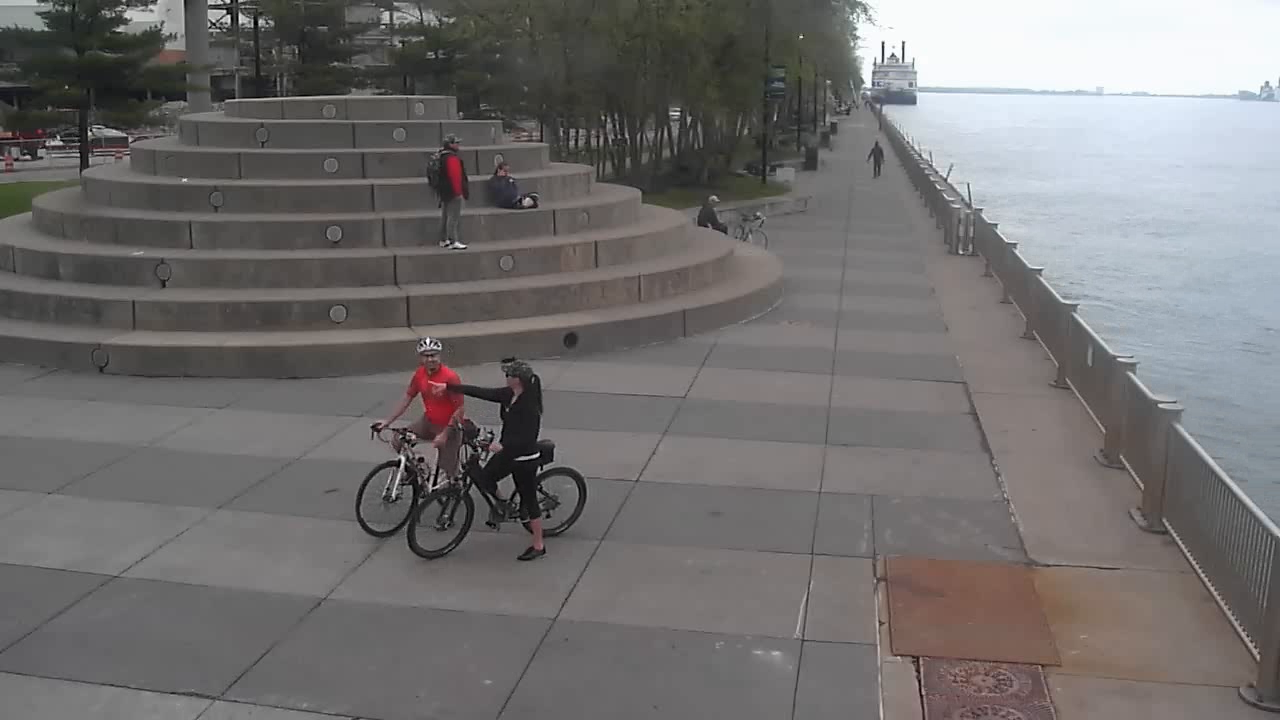}
    \vspace{-1.5em}   % make gap between caption and text smaller
    \caption{Sitting on steps}    
    \label{fig:dc_det_sparse}
    \end{subfigure}
    \,
    \begin{subfigure}[b]{0.23\textwidth}  
    \centering 
    \includegraphics[width=\textwidth]{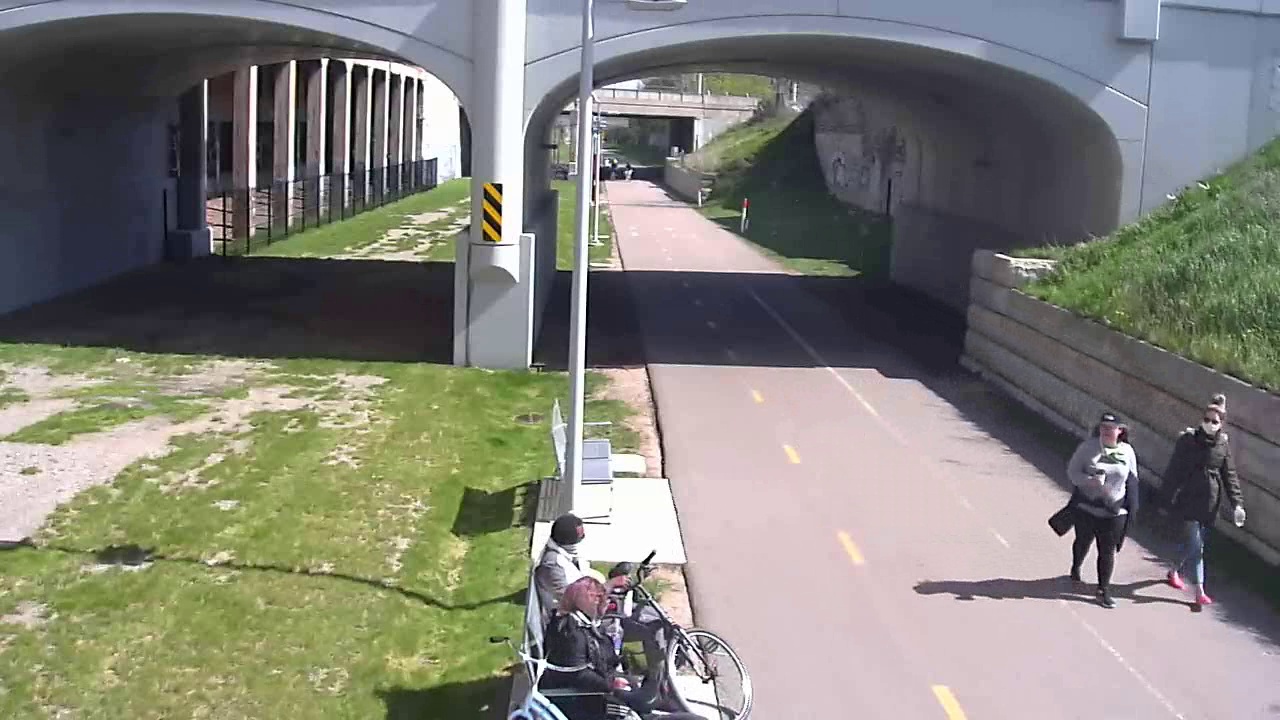}
    \vspace{-1.5em}   % make gap between caption and text smaller
    \caption{Sitting and walking in pairs}    
    \label{fig:dc_det_dense}
    \end{subfigure}
    \\
\caption{Video frames with people taking activities including (a) fishing and bicycling, (b) talking between friends, (c) sitting on steps, and (d) walking in pairs at the Detroit Riverfront Parks managed by Detroit Riverfront Conservancy (DRFC) during COVID-19 (May 2020).}
\vspace{-1.5em}   % make gap between caption and
\label{fig:intro_drfc}
\end{figure}

The spread of coronavirus disease 2019 (COVID-19), an infectious disease caused by severe acute respiratory syndrome coronavirus 2, has taken the world into pandemic in only a few months.The World Health Organization (WHO) declared the COVID‑19 outbreak a public health emergency of international concern (PHEIC) on Jan.30 2020 and a pandemic on Mar.11 2020 \cite{world2020who}. On Feb.26 2020, US CDC Confirms Possible Instance of Community Spread of COVID-19 in California and U.S. \cite{CDC2020spread} The US federal government and state governments are making efforts to warn people, to prepare health systems, and to suppress the spread of the pandemic.  For example, the State of Michigan declared a state of emergency on Mar. 10 2020 \cite{michigan2020spread} and the \textit{stay-at-home} executive order (2020-42 (COVID-19)) \cite{michigan2020home} on Apr.9 2020. Together with the these legislative efforts, US Centers for Disease Control and Prevention (CDC) recommends \textit{social distancing} and other hygienic practices, such as avoiding physical contact with hard surfaces and wearing protective facial coverings or masks. According to CDC guidelines \cite{CDC2020social}, \textit{social distancing} (a.k.a. physical distancing) means "keeping space between yourself and other people outside of your home".  The practice of social distancing \cite{CDC2020social} includes: (1) staying at least 6 feet (about 2 arms' length) from other people, (2) abstaining from gathering in groups, (3) staying out of crowded places and avoiding mass gatherings.  CDC also suggests wearing facial coverings or masks in public settings for protection due to a significant portion of asymptomatic and pre-symptomatic individuals \cite{CDC2020mask}.

\begin{figure} [tp]
\vspace{-1.0em}   % make gap between caption and
    \centering
    \includegraphics[width=1.0\linewidth]{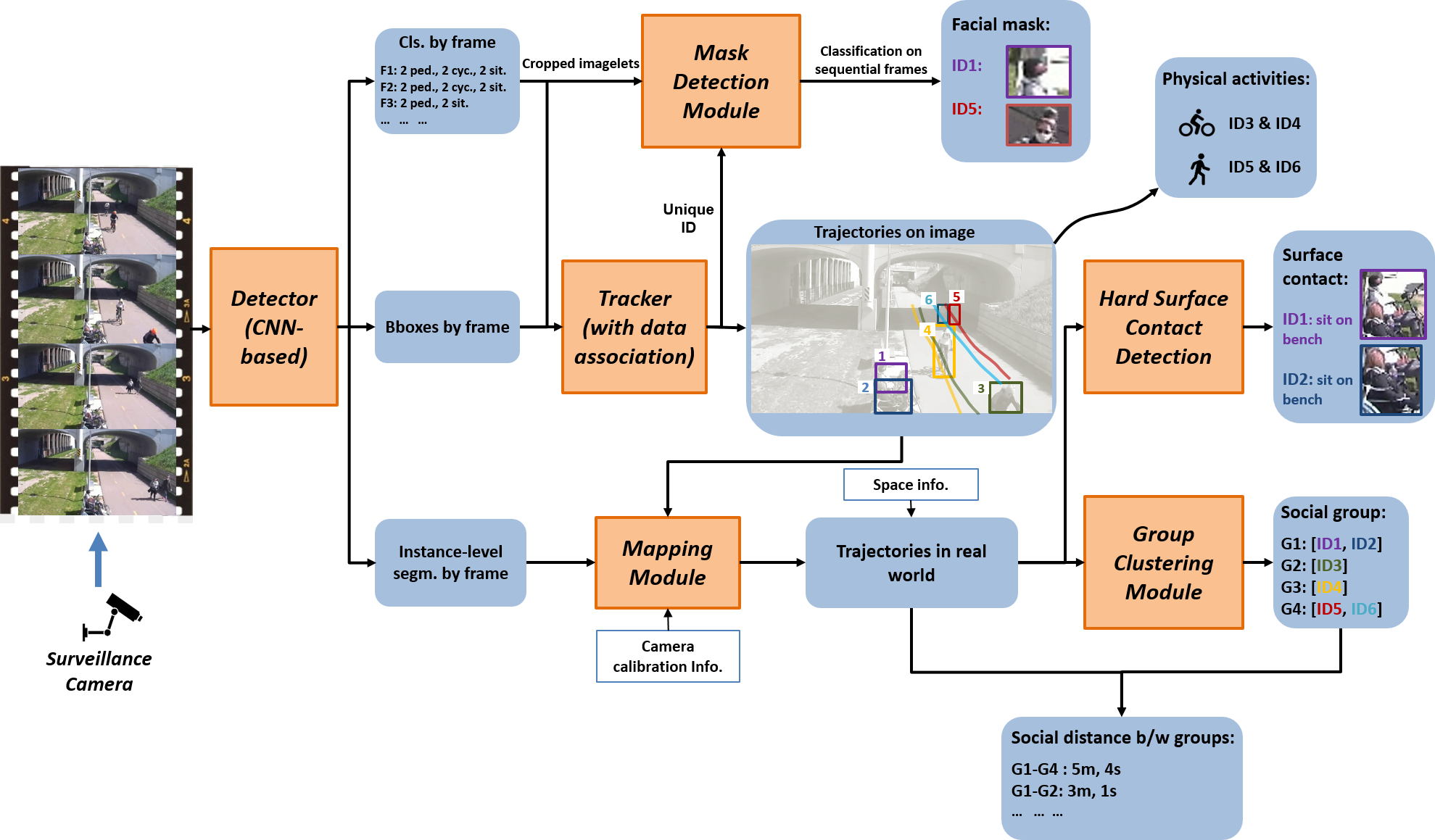}
    \caption{Architecture of the multi-task, automatic, sensing framework to explore relationships of user-user and user-facilities.}
\vspace{-1.5em}   % make gap between caption and
\label{fig:teaser_frame}
\end{figure}

Public open spaces (POS) provide important platforms for residents to perform physical activities which can reduce the risk of chronic disease \cite{francis2012quality}. Due to the pandemic and the executive orders, many people have to stay and work at home for extended periods of time. Additionally, gyms, recreational centers, and non-essential retail stores have been temporarily closed to the public. This has led to a sharp increase in the use of POS as people are seeking opportunities for physical exercise or simply needing to spend more time outdoors in a natural environment as shown in Fig.~\ref{fig:intro_drfc}. During this time public space managers are faced with a challenging question: how to maintain the utility of the POS and at the same time ensure the safety of both park patrons and workers? It is not economically efficient or safe for park employees to manually monitor (large) POS and possibly intervene to correct poor hygienic practices. Hence, there is an urgent need to monitor social distancing and hygienic practices of patrons utilizing and enjoying park facilities within POS. 

The objective of the paper is to propose an autonomous sensing approach (as shown in Fig.~\ref{fig:teaser_frame}) to recognize patrons' physical activities in POS using pre-installed surveillance cameras and to monitor the practices of social distancing and hygienic protections which suppress the wide spread of COVID-19. The contribution of this proof-of-concept work lies in: (1) it proposes an efficient and safe tool to inform POS managers or government of the POS utilization with healthy precaution during the pandemic; (2) it is cost efficient for reusing the monocular cameras (which have already been pre-installed for surveillance purposes) for social sensing tasks; (3) multi-task capacities is incorporated on the sensing framework, spatio-temporal and social data could be collected for future applications (e.g. urban design or improvement of parks and facilities). The cameras in current use at the Detroit riverfront lack sufficient resolution for facial recognition. To further ensure the privacy, the sensing framework is designed that only anonymized version of the patrons information is extracted with raw images not saved once the system is fully trained. 

The structure of the paper is as follows. First, relative studies (e.g. human detection and tracking, crowd analysis and group detection, etc.) will be presented. Second, algorithms of different function modules (e.g. detection, tracking, activity mapping, group detection, mask detection, etc.) with tempo-spatial assessment of social dynamics will be described. Third, the evaluation of each function module and experiment results will be represented as well as the applications (e.g. activity mapping, measurement of social distance measuring, patron-facility contact detection, etc.). In the end, the concluding remarks will be summarized.

\section{Related Works} \label{sec:related}
\textbf{Human Detection and Tracking}
Object detection models are utilized to identify and locate objects in images. Region-based detection models (e.g. Fast R-CNN \cite{girshick2015fast} and Faster R-CNN \cite{ren2015faster}) rely on region proposal networks (RPN) \cite{ren2015faster} and convolutional neural networks (CNN) to estimate bounding boxes (bboxes) of objects. In contrast, single-stage models (e.g. YOLO \cite{redmon2016yolo} and SSD \cite{liu2016ssd}) perform object detection without a separate region proposal step.  Although the former methods suffer from comparatively slow detection speed, they outperform the latter in detection accuracy \cite{zhao2019ObjDetReview}.  Mask R-CNN \cite{he2017mask} is a region-based detection method that yields richer information of detected objects by providing instance segmentation and bbox coordinates. Furthermore, detected contours can provide location information of specific body parts \cite{li2017humanParsing}. Recently, new anchor-free detectors (e.g. FCOS \cite{tian2019fcos}, FoveaBox \cite{kong2019foveabox}) have been developed to achieve higher performance in detecting accurate bboxes without using anchor references. The tracking-by-detection paradigm is usually adopted to perform due to the good performance of the aforementioned, emerging CNN-based detectors. For example, SORT \cite{Bewley2016sort} and DeepSort \cite{wojke2017deepsort} are widely used robust online tracking algorithms with detector-tracker coupled schemes.  Tracking is accomplished in a general tracking scenario where videos or video frames have not been rectified and no prior ego-motion information is provided. In this study, the DeepSort algorithm \cite{wojke2017deepsort} is adopted to build the tracking module and includes various components (e.g. Kalman filter, cascade matching, IoU assignment \cite{Bewley2016sort}, etc.). The tracker will perform data association tasks based on frame-based detection results from CNN detectors.

\textbf{Human Activity in Public Open Spaces}  
Manually observing social interactions with the naked eye, either in real-time or though recorded videos, has been the primary method for studying social behavior \cite{whyte1980social}. Researchers have been working on human activity recognition (HAR) using different sensors, including non-vision based (e.g. wearable) and vision-based sensors. In \cite{lara2012wearable, yang2008distributed}, multiple wearable sensors (e.g. accelerometers, gyroscopes, and magnetometers) are attached to the body of a subject to measure motion attributes to recognize different activities. However, wearable sensors are intrusive to users \cite{jalal2017depthvideo} and can only cover a very limited number of users in a POS. Traditional CV methods using vision-based sensors usually rely on a few visual features (e.g. HoG, local binary pattern, or RGB), which can not achieve robust pedestrian detection (especially under extreme illumination conditions). \cite{yan2005public} is one of the few studies that employed a computer vision-based method to measure human activity (e.g. sitting or walking) in a POS. The people detection method was based on background subtraction and blob detection.  In the past few years, deep learning \cite{lecun2015deep, goodfellow2016deep} methods using deep neural networks have grown rapidly and are drawing attention as the result of their supreme performance in many applications.  Others have used deep features of images to extract high-level representation for activity recognition tasks \cite{caba2015activitynet,gkioxari2018interaction}. A complete HAR study \cite{vrigkas2015HARreview} usually consists of two steps: 1) person detection, and 2) activity/body shape recognition based on feature representations (e.g. silhouette representation). For simplicity, the recognition of different activities will be embedded within the classification task of the instance segmentation using Mask R-CNN detector in this study.

\textbf{Collective Motion and Small Groups}
Groups are defined as clusters of people with similar features (e.g. location, speed, moving direction). Understanding group-level dynamics and properties is important for a large range of applications. Group dynamics have been extensively studied in sociopsychological and biological research as primary processes influencing crowd behaviors (e.g. animal behavior in crowd). Researchers in computer vision \cite{shao2014scene} used video surveillance feeds to study internal intra-group dynamics (e.g., collecvtiveness, stability, and uniformity) and external inter-group dynamics (e.g.,conflicted movement between passing groups). Collective behavior and the collective motion of groups are attractive phenomena in both nature and human society. In social psychology literature, collective behavior is referred as a generic term for the often extraordinary and dramatic actions of groups and the individuals within \cite{lindzey1954handbook}.
Models of collective behavior fall between two extremes. At one extreme entire crowds are considered as one entity with a homogeneous "group mind" \cite{lindzey1954handbook}. While the other extreme treats individuals as independent units acting to maximize their own utility and making local decisions based on the principle of least effort \cite{still2000crowd}. The real-world scenario falls between the two extremes, both isolated individuals and small groups of people coexist within a crowd \cite{ge2012vision}.  Lau et. al. \cite{lau2010multi} is one of the first researchers to cluster measured 3D data points (from a laser range finder) into groups of human-size blobs for merging and splitting groups over time based on \textit{proxemics} \cite{hall1963system}.  Proxemics can be used to define groups based on ranges of personal and social interaction distances - though if only location information is considered this method is not robust. For example, if an isolated individual approaches near a group and gets within a certain distance they will be treated as a member of the group.  Models describing human traffic flow can either be macroscopic, focusing on space allocation, or microscopic focusing on pedestrian groups and interactions between agents \cite{may1990traffic}.In this paper, the authors will focus on micro-level traffic flow and individual tracking. One objective is to analyze sets of trajectories or tracklets to identify pedestrian groups within crowded environments. Identifying groups is necessary for the social distancing aspect of this study.

\textbf{Facial Occlusion/Mask Detection:} 
While researchers have worked towards masked face detection \cite{ghiasi2014occlusion,opitz2016grid,yang2015facial}, Ge et. al. \cite{ge2017detecting} presented the largest and richest masked face dataset (MAFA). The MAFA dataset includes 30,811 images consisting of 35,806 masked faces. MAFA also provides detailed annotations describing face and mask location as well as mask type and degree of occlusion. Ge et. al. \cite{ge2017detecting} proposed a three-module framework which utilized LLE-CNNs for masked face detection. The methodology in \cite{ge2017detecting} was able to outperform 6 other state-of-art face detectors in detecting occluded faces from the MAFA dataset by over 15\%, including the hierarchal deformable part model presented in \cite{ghiasi2014occlusion}. In this paper, a custom dataset will be presented and used for training a CNN-based facial mask detector using in-field surveillance videos (with relative low resolution).

\section{Method} \label{sec:method}
\subsection{Patron Recognition and Activity Mapping}
\subsubsection{Detection on Video Frames}
In the proposed sensing framework, the detection module adopts Mask R-CNN \cite{he2017mask} which provides instance segmentation which helps achieve more accurate mapping. The architecture of the Mask R-CNN model includes a convolutional neural network (CNN) backbone structure (e.g. ResNet \cite{he2016deep}), a region proposal network (RPN) \cite{ren2015fasterrcnn}, a ROI Align module \cite{he2017mask}, and a classifier (e.g. to classify and regress bboxes and masks). In Mask R-CNN the bbox regression process is performed in two cascaded stages: the first stage is the objectiveness (class-agnostic) bbox regression, which is processed in the RPN, and the second stage is the class-based bbox regression which is processed after the regions of interest are generated. This study utilizes ResNet \cite{he2016deep} as the CNN backbone to extract feature maps from the input images. A feature Pyramid Network (FPN) \cite{lin2017FPN} is also implemented with the CNN backbone to extract multi-scale features (in both semantic and location aspects)  with the bottom-up top-down approach.

This paper will use Mask R-CNN as the detector in the proposed sensing framework and use a single-stage detector just for comparison. The training process, dataset selection, and evaluation will be presented in Section. \ref{sec:experiment}. 

\subsubsection{Tracking by Detection} \label{sec:method_tracking}
\begin{figure}
\vspace{-1.0em}   % make gap between caption and
\centering
\includegraphics[width=1.0\linewidth]{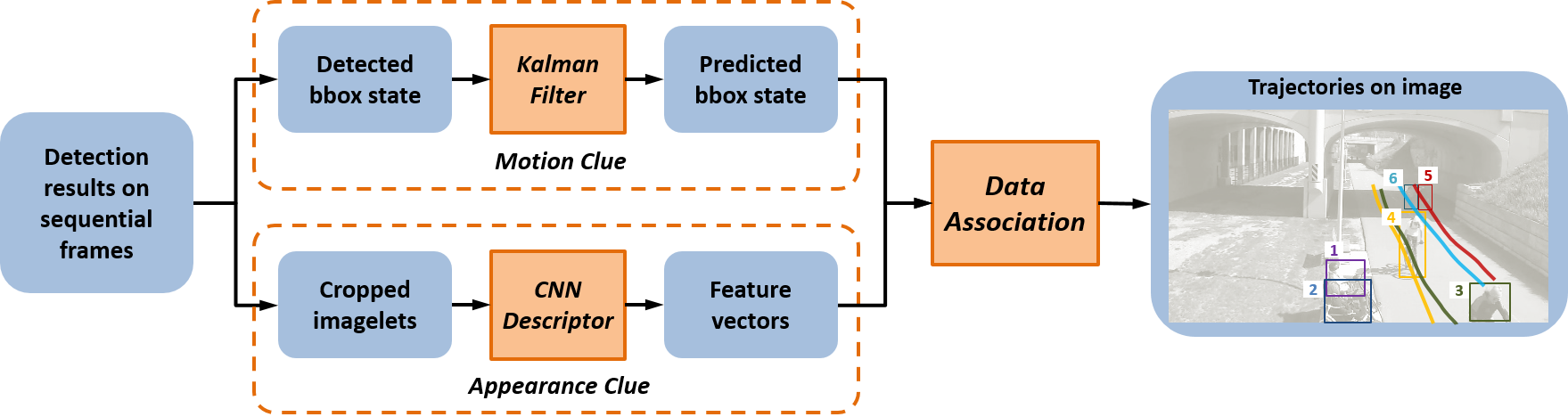}
\caption{Schematic of the tracking module using detection results on sequential frames.}
\vspace{-1.5em}   % make gap between caption and
\label{fig:method_tracker}
\end{figure}

In this study, a popular tracking paradigm, tracking-by-detection, is adopted to build the patron tracking and unique ID assignment. Deep Sort algorithm \cite{wojke2017deepsort} is utilized to build the tracking module integrated by various components (e.g., Kalman filter, cascade matching, IoU assignment, etc.). The tracking problem is formatted as an assignment problem between existing tracks (based on detection and tracking results on previous frames) and new detection results on the current frame. The proposed detection module serves as the detector while the Hungarian algorithm \cite{kuhn1955hungarian} solves the assignment problem. The tracking algorithm utilizes both motion and appearance for data association metrics as shown in Fig.~\ref{fig:method_tracker}. It has a more robust tracking performance (especially under occlusion condition) than other tracking methods (e.g. SORT \cite{Bewley2016sort} ). 

A wide residual network consisting of ten convolution layers \cite{wojke2017deepsort} is used to extract the appearance features of cropped images from the original image (within detection bboxes). The CNN weights have been trained on a large-scale Re-ID dataset \cite{zheng2016mars}. The CNN \cite{zagoruyko2016wide} obtains discriminative contexts from the cropped images (of detected patrons) and maps them onto vectors with fixed length. After feature extraction using the CNN, the discriminative feature vectors of two different patrons are expected to vary widely, while discrimintive feature vectors of a single patron taken from different frames are expected to be similar.

The tracking module (as shown in Fig.~\ref{fig:method_tracker}) utilizes both motion related and appearance related information of the detected/tracked objects. The tracking procedure is based on the detection results (in a frame-by-frame fashion) that are obtained using the detection module. The motion related information is updated by the Kalman filter algorithm \cite{kalman1960KalmanFilter} and appearance related information is generated by a Re-ID CNN. Both the distances in motion ($d_{mot}$ in squared Mahalanobis distance) and in appearance ($d_{app}$ in cosine distance) between a detection and an existing tracker are used in the following data association step. Data association between the detected results (on current frame) and the identified tracks (on previous frames) is performed across sequential frames. Trajectory results of patrons are obtained after the data association process.

\subsubsection{Mapping using Monocular Camera} %\label{sec:method_mapping}
A pinhole camera model \cite{ma2012imagemodel} is used to develop a mapping module to relate detected results on images to the real world. In the pinhole camera model for monocular cameras (e.g. most surveillance cameras), a 3D point location in the world coordinate system (WCS) $\{X, Y, Z\}$ can be projected onto the image plane with location in the 2D pixel coordinate system (PCS) $\{u, v\}$.  The projection can be achieved by using a perspective transformation:

\begin{equation}
	s\bm{m'}=\bm{A \left[ R|t \right] M}
\label{eq:perspective_transformation}
\end{equation}

\noindent where $s$ is the scaling factor, $\bm{A}$ is the intrinsic matrix, $\bm{R}$ is the rotation matrix, $\bm{t}$ is the translation vector represented in the camera coordinate system (CCS), $\bm{m}=\left[ u,v,1\right]^{T}$, and $\bm{M}=\left[X,Y,Z,1\right]^{T}$.

\subsection{Social Distance and Group Clustering}

\begin{figure}[bp]
\vspace{-1.0em}   % make gap between caption and text
\centering
    \begin{subfigure}[b]{0.14\textwidth}
    \centering
    \includegraphics[width=\textwidth]{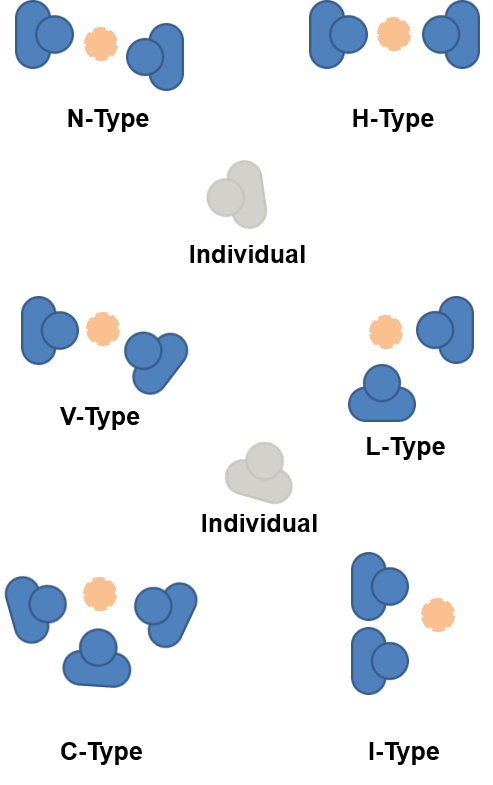}
    \vspace{-1.5em}   % make gap between caption and text smaller
    \caption{} 
    \label{fig:method_group_conversation}
    \end{subfigure}
    \hspace{0.05\textwidth}
    \begin{subfigure}[b]{0.22\textwidth}  
    \centering 
    \includegraphics[width=\textwidth]{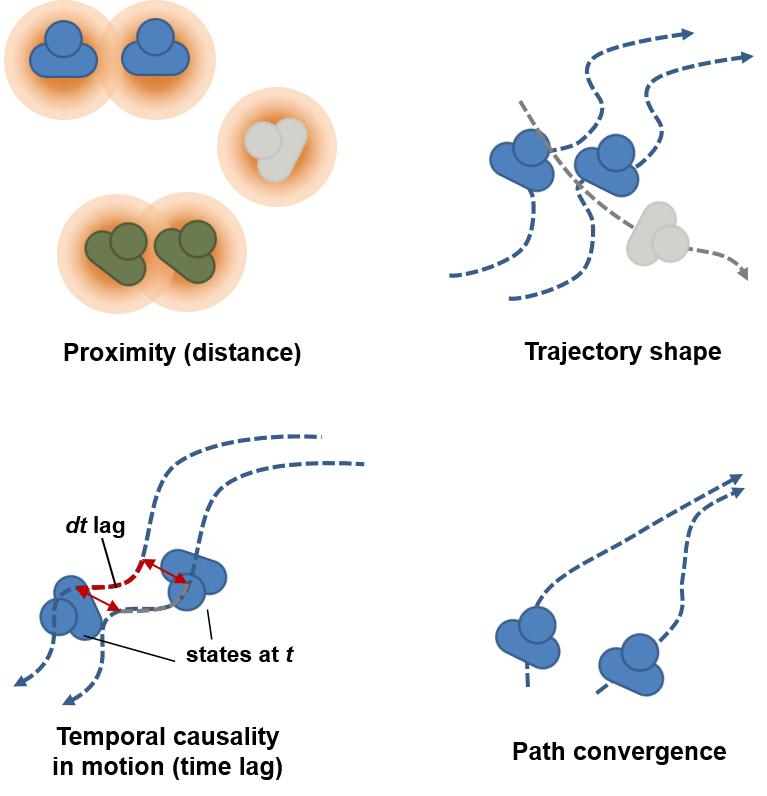}
    \vspace{-1.5em}   % make gap between caption and text smaller
    \caption{}    
    \label{fig:method_group_motion}
    \end{subfigure}
    \\
\vspace{-0.2em}   % make gap between caption and
\caption{Schematics of (a) types of F-formation of conversational groups according to the spatial layout, and (b) relation features that can be extracted for group in motion. Note the orange circle represents the o-space in F-formation.}
\vspace{-1.5em}   % make gap between caption and
\label{fig:method_features}
\end{figure}

Small groups are more prevalent than large crowds in public spaces (e.g. pedestrian scenes, cyclist scenes). This study focuses on small groups, which are typically composed of family members and friends and are much more common at public community parks than heavy crowds composed of tens or hundreds of individuals. Properly identifying these small groups is especially relevant in the context of social distancing. Members within a group (e.g. a family with children or a couple) have clear awareness of the travel history and health of their companions and so it is less imperative and slightly unrealistic to expect these individuals to maintain social distancing between others in the group. However, due to infection dangers from long-time exposure to unfamiliar individuals, it is still imperative for groups to maintain social distancing between other groups and individuals. Therefore, in order to adequately monitor social distancing in POS, it is necessary to firstly identify and distinguish different groups.

In POS, groups are usually either in conversation or in (collective) motion. Although conversational groups can be dynamic, they tend to change slowly compared to groups in motion. Conversational groups can be described using various F-type formations \cite{ciolek1980environment} and physical location with pose orientation of individuals can be used to infer possible o-space and memberships (Fig.~\ref{fig:method_group_conversation}). Recognition methods for both types of groups are being actively researched and developed. 
However, due to the usage patterns of the site (DRFC) studied, this paper focuses on the recognition of groups in motion (Fig.~\ref{fig:method_group_motion}). The DRFC is primarily used by patrons for physical activities (e.g. walking, jogger, and bicycling) on a stroll, especially during the pandemic period. This study aims to recognize social groups in motion by using physical locations and other social cues derived from information captured by the monocular surveillance cameras.

\subsubsection{Group Detection Method}
The group detection task is treated as a clustering problem, in which patrons detected in current time window $\CMcal{T}_{k}$ compose the set $\CMcal{M} = \{i,j,...\}$. $\CMcal{Y}(\CMcal{M})$ is defined a set of all possible partitions of $\CMcal{M}$. The task is to find the best possible partition $y \in \CMcal{Y}(\CMcal{M})$ that satisfies a preset criteria. The partition problem is solved using Correlation Clustering (CC) algorithm \cite{bansal2004correlation} which is expressed as:
\begin{equation}
   CC = \argmax_{\bm{y} \in \CMcal{Y}(\CMcal{M})}  \sum_{y \in \bm{y}}\sum_{i\neq j \in y} (W_{\bm{f}})_{ij}
\end{equation}
\noindent where $W_{f}$ is an affinity matrix based on the feature selection and $(W_{f})_{ij}$ is the element corresponding to $i$-th row and $j$-th column. If $(W_{f})_{ij} >0$, $i$ and $j$ belong to the same group and vice versa. $|(W_{f})_{ij}|$ represents the confidence of prediction or the predicted pairwise affinity.

The affinity between two elements (e.g. $i$ and $j$) is a weighted sum of the similarity features.
\begin{equation}
   (W_{\bm{f}})_{ij} = \bm{\alpha}^{T}(\bm{1}-\bm{f}(i,j)) - \bm{\beta}^{T}\bm{f}(i,j)
\label{eq:para_feature}
\end{equation}
\noindent where $\bm{\alpha}$ and $\bm{\beta}$ are parameters that can be either selected or trained using a large dataset with ground-truth group clusters.

\begin{figure}[tp]
 \vspace{-1.0em}   % make gap between caption and
\centering
\includegraphics[width=1.0\linewidth]{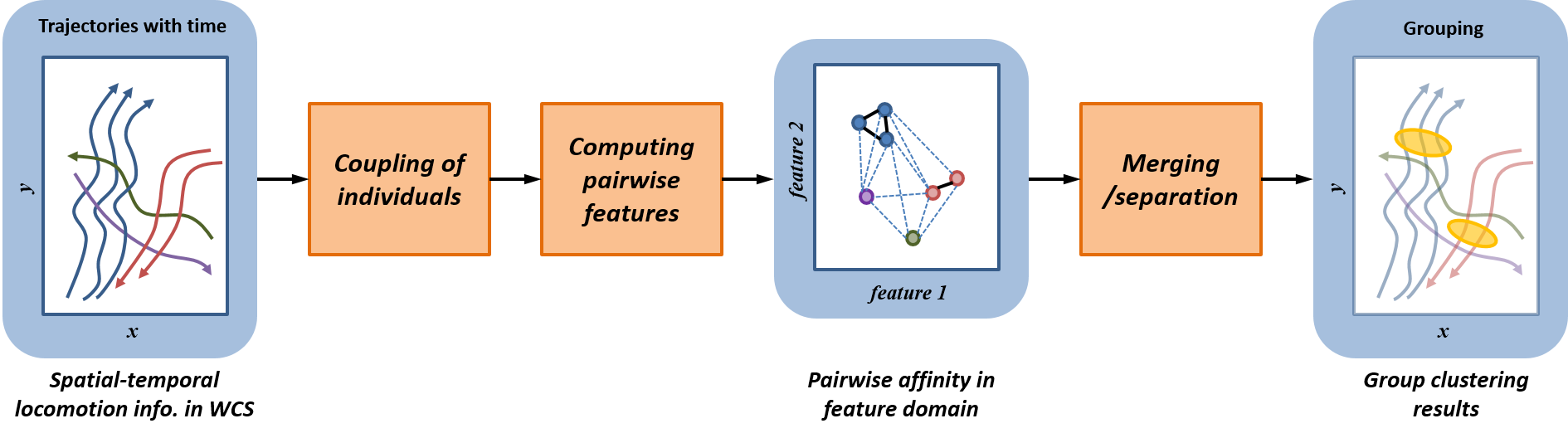}
\caption{Schematic of group detection using spatial-temporal information of park users.}
\vspace{-1.5em}   % make gap between caption and
\label{fig:method_grouping}
\end{figure}

As shown in Fig.~\ref{fig:method_grouping}, the pairwise features or feature vectors (depending on the number of features adopted) are calculated for all possible couples (e.g. $(i,j)$) within $\CMcal{M}$. The pairwise features are then multiplied by the parameters to get a weighted sum representing the affinity within each couple. The Cluster Correlation algorithm is then employed to decide whether to merge couples or separate individuals from couples. $y_{opt}$ is an optimized partition of $\CMcal{M}$ representing the final result of the group detection module.

\subsubsection{Similarity Features for Group Members}
This study will explore the features or coherence from any two individuals and then determine the relationship of the two by using some relation features. The most straightforward thoughts is the spatial information of patrons in POS.  Social science researchers \cite{mcphail1982using} used cascaded metrics (all three need to be satisfied) to determine the group membership of any two individuals: (1) the location of the two are within 7 feet of each other (e.g., $s_i(t)-s_j(t)<\tau_s$); (2) the speed difference is within 0.5 ft/sec (e.g., $v_i(t)-v_j(t)<\tau_v$); and (3) the directions of motion are within 3 degrees of each other. Inspired by the metrics for determining group membership between two individuals, Ge et. al. \cite{ge2012vision} built thresholds (e.g. distance and speed difference) to filter out unqualified couples and designed a relatively robust feature which uses the combination of location/proximity and velocity cues for group membership reasoning. Symmetric Hausdorf distance was adopted for measuring inter-group closeness. The temporal information is only considered when checking the affinity or feature across nearby video frames.

Apart from location and velocity, Solera et. al. \cite{solera2015socially} extends to extract additional social features from trajectories, such as motion causality, trajectory shape similarity, and common goals. The aforementioned studies used feature based tracking method (e.g. corner features of people) while this study uses (CNN) detection based tracking method. The affinity features will be extracted from the spatial-temporal information of the patrons in POS environment.These features can be used to form an affinity distance of the weighted combination for the following optimization in group clustering. The parameters for the weighted combination can be obtained from a structural support vector machine that is trained on grouping dataset. A pairwise feature vector \cite{solera2015socially} within time window $\CMcal{T}_{k}$ (usually a fixed interval of several seconds) for each couple is expressed as:

\begin{equation}
    f_\text{traj}^{k}(i,j) = \left[ f_1(i,j), f_2(i,j), f_3(i,j),f_4(i,j) \right]
\end{equation}

\noindent where the $f_\text{traj}^{k}(i,j)$ is composed of four features of the $(i,j)$ pair, including proxemics \cite{hall1966hidden}, trajectory shape similarity \cite{berndt1994dtw}, motion causality \cite{granger1969investigating}, and path convergence \cite{lin2013heat} during time window $t \in \CMcal{T}_{k}$.

\subsection{Inferring People-Facility Contact and Mask Wearing}

\begin{figure}[bp]
\vspace{-1.2em}   % make gap between caption and
    \begin{subfigure}[b]{0.23\textwidth}
    \centering
    \includegraphics[width=\textwidth]{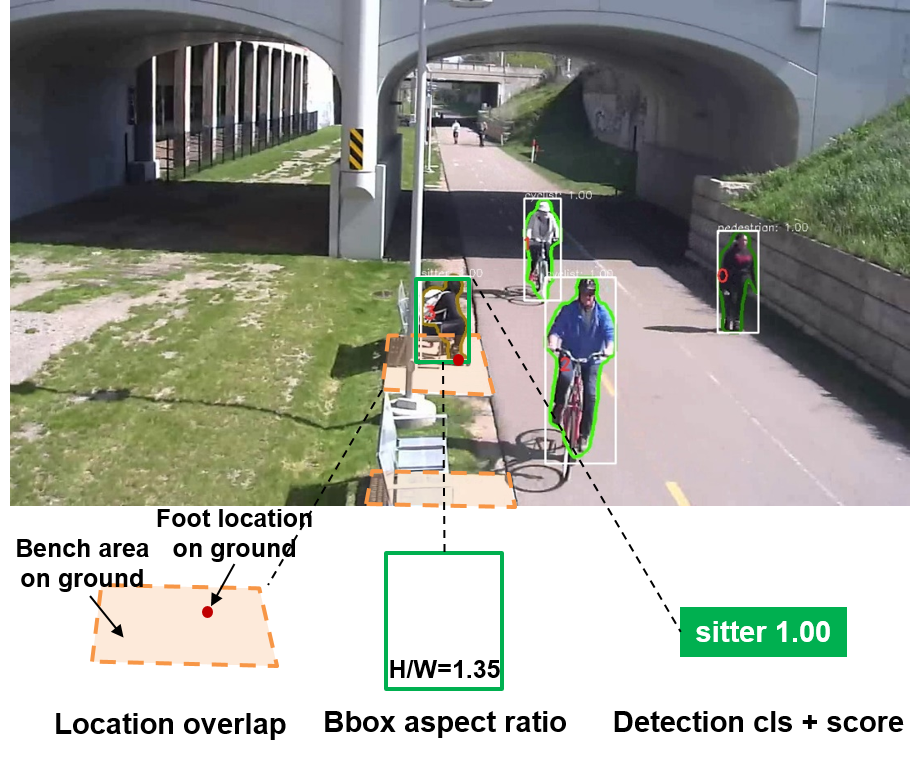}
    \vspace{-1.5em}   % make gap between caption and text smaller
    \caption{Sitting on benches} 
    \label{fig:method_contact_sit}
    \end{subfigure}
    \hspace{0.005\textwidth}
    \begin{subfigure}[b]{0.23\textwidth}  
    \centering 
    \includegraphics[width=\textwidth]{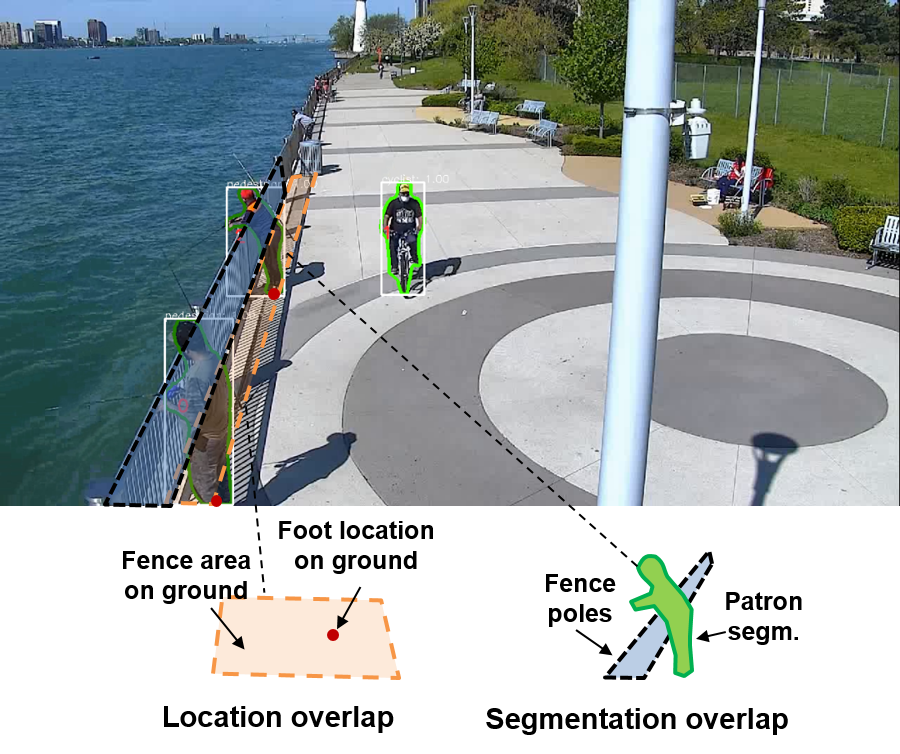}
    \vspace{-1.5em}   % make gap between caption and text smaller
    \caption{Leaning on fence poles}    
    \label{fig:method_contact_lean}
    \end{subfigure}
    \\
\vspace{-1.0em}   % make gap between caption and
\caption{Schematics of inferring contact with hard surfaces on (a) benches and (b) fence poles in POS.}
\vspace{-1.5em}   % make gap between caption and
\label{fig:method_contact}
\end{figure}

The COVID-19 virus can reportedly stay active on hard surfaces for extended periods of time, surviving many hours past the initial contact time. Gloomy weather, humidity, and low temperatures can extend the COVID-19 virus hard surface survival time even longer. Thus the physical contact between patrons and hard surfaces in POS is important to track although can be challenging to infer from surveillance footage (2D images) alone. However, with some prior knowledge of the environmental setting, geographical information of potentially hazardous surfaces (e.g., park facilities and architectural designs) can be incorporated into the proposed sensing method, and it can be feasible to quickly identify the usage of park facilities. The physical contact between patrons and park facilities can be inferred from the relationship of detected/tracked patrons with geospatial information (e.g., location of chairs, trashcans, and fencing). For example, if a person sits on a bench (as shown in Fig.~\ref{fig:method_contact_sit}) or leans against a fence (Fig.~\ref{fig:method_contact_lean}), the event can be inferred, recorded, and timed with a tracking module. Although possible interventions are not studied in this paper, POS owners or managers can take advantage of the monitoring results and revise cleaning strategies or implement further signage warning patrons of hygienic risk.

\begin{figure}[tp]
\vspace{-1.0em}   % make gap between caption and
\centering
\includegraphics[width=0.9\linewidth]{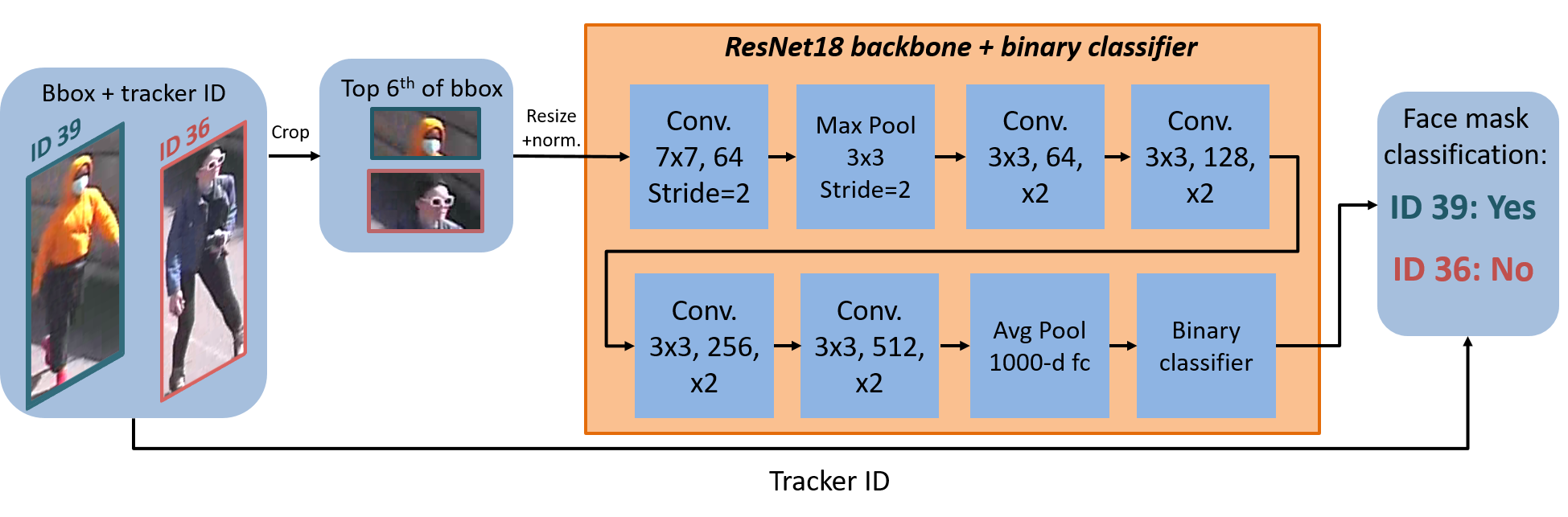}
\caption{Schematic of mask detection process using CNN-based classifier.}
\vspace{-1.5em}   % make gap between caption and
\label{fig:method_mask}
\end{figure}

The CDC recommends wearing face masks in public settings to help slow the spread of COVID-19. In order to track pedestrian use of facial coverings a mask detection tool was developed. The mask detection tool utilizes the detected pedestrian bboxes from the Mask R-CNN detection model as well as tracking information associated with the detection. If a tracked pedestrian is determined by the tracking state information to be heading towards the camera, the head area (top $6^\text{th}$) of the bbox is cropped and fed into the mask detection tool. For better performance the face mask detection tool will wait until a patron is closer to the camera before initializing, bboxes with insufficient data (when pedestrians are farther away from the camera) are ignored. Surveillance cameras in POS are usually low resolution and are installed higher up on light poles for a larger field of view. For these reasons face mask detection can be challenging. For example, when patrons are near the surveillance cameras used in this study the cropped head areas are on average only 40 $\times$ 80 px$^2$, with the defining feature (face mask) typically being between 10-20 $\times$ 10-20 px$^2$. The face mask detection tool utilizes a CNN-based binary classifier with a ResNet-18 backbone (as shown in Fig.~\ref{fig:method_mask}) to classify the cropped head areas and associated tracker IDs as having a facial covering or not. In order to ensure the robustness in detection, the top 1/6 of each image is further cropped, normalized (by color), and resized into a size of 64 $\times$ 64 px$^{2}$ before passing through the CNN-based classifier.

\section{Experiment and Result} \label{sec:experiment}
\subsection{Activity Recognition in OPOS}

\begin{table} [bp]
\vspace{-1.2em}   % make gap between caption and
\caption{Performance of Mask R-CNN and RetinaNet evaluated on the OPOS testing dataset (unit: \%).}
\label{tab:experiment_detection}
\footnotesize  % size of the font
\centering
\begin{tabular}{>{\centering\arraybackslash}p{1.5cm}>{\centering\arraybackslash}p{0.4cm}>{\centering\arraybackslash}p{0.4cm}>{\centering\arraybackslash}p{0.4cm}>{\centering\arraybackslash}p{0.4cm}>{\centering\arraybackslash}p{0.4cm}>{\centering\arraybackslash}p{0.5cm}>{\centering\arraybackslash}p{0.8cm}}
\toprule
\multirow{2}{*}{Detector}                                                             & \multicolumn{6}{c}{AP per ppl. class}                                                                   & \multirow{2}{*}{mAP (ppl.)} \\
                                                                                                          & \textit{ped.} & \textit{cycl.} & \textit{scoot.} & \textit{skat.} & \textit{sitter} & \textit{ppl.oth} &                                  \\
\midrule
mask (bbox)                                                        & 96.36       & 96.50        & 89.39         & 89.52        & 89.14         & 74.08             & \textbf{89.17}                          \\
mask (segm)                                          & 96.31       & 96.46        & 89.39         & 89.52        & 89.52         & 73.11             & \textbf{89.05}   \\
retina (bbox)                                         & 97.59       & 98.10        & 95.53         & 38.05        & 89.89         & 68.43              & 81.26   \\ \bottomrule                      
\end{tabular}
\vspace{-1.8em}   % make gap between caption and
\end{table}
\normalsize

Among the existing models, Mask R-CNN has a much better trade-off balance between accuracy and speed and is a good candidate for the detector in the sensing framework. However, for comparison, this work also presents the detection results from one-stage detectors in the POS environment. The weights of the CNN backbone in the detection models are firstly trained on the ImageNet-1K dataset \cite{deng2009imagenet} and COCO\_2017 dataset \cite{lin2014microsoftcoco}. Transfer learning for Mask R-CNN is performed using maskrcnn-benchmark platform \cite{massa2018mrcnn}. The mini batch size is set as 2 images/batch and horizontal flipping data augmentation is adopted for training.  The schedule includes 90k iterations and starts at a learning rate of 0.0025.  The learning rate is decreased by a factor of 0.1 after 60k and 80k iterations and finally terminates at 90k iterations. This schedule results in 25.56 epochs over the OPOS\_training dataset \cite{sun2020wacv} which consists of 7043 images with 16902 annotated bboxes and (instance) segmentations. The OPOS dataset include classes of people in POS with different physical activities (e.g. bicycling, scootering, sitting, etc.) and classes of usual objects (e.g. cars, strollers, dogs, etc.). In order to ensure a thorough training, the order of the images in the training dataset is shuffled after each epoch. To compare the performance of the trained detectors, evaluation is performed adopting AP50 metrics on OPOS testing dataset with 783 images with 2000 annotated objects. As shown Table~\ref{tab:experiment_detection}, the Mask R-CNN outperforms RetinaNet by 7.91\% in overall people detection (mean average precision in bbox). Mask R-CNN also provides segmentation information that enables correct localization of patrons compared to bbox.

\begin{table*}[tp]
\vspace{-1.2em}   % make gap between caption and
\caption{Evaluation of the tracking performance (detector and tracker with same parameter setting) on a custom tracking dataset using MOT-challenge metric and total unique ID count (ID Ct.).}
\centering
\label{tab:experiment_tracker}
\footnotesize  % size of the font
\begin{tabular}{clcccccccccc}
\toprule
\multicolumn{2}{c}{Method}                                                                                 & MOTA    & MOTP    & Prcn    & Rcll    & GT & MT & PT & ML & IDs & ID Ct. \\ \midrule
\multirow{3}{*}{\begin{tabular}[c]{@{}c@{}}detect+track   \\ (raw results)\end{tabular}}     & $n_{skip}=1$  & 80.30\% & 87.60\% & 95.00\% & 85.70\% & 53 & 33 & 20 & 0  & 36 & 59 \\
                                                                                             & $n_{skip}=2$  & 72.40\% & 85.70\% & 95.30\% & 77.70\% & 53 & 18 & 35 & 0  & 33 & 59  \\
                                                                                             & $n_{skip}=3$  & 56.40\% & 83.80\% & 92.80\% & 63.00\% & 53 & 6  & 42 & 5  & 28 & 52 \\
\midrule
\multirow{3}{*}{\begin{tabular}[c]{@{}c@{}}detect+track  \\ (filtered results)\end{tabular}} &$n_{skip}=1$ & 80.40\% & 87.60\% & 95.00\% & 85.70\% & 53 & 33 & 20 & 0  & 36  & 57\\
                                                                                             & $n_{skip}=2$ & 72.60\% & 85.70\% & 95.50\% & 77.60\% & 53 & 18 & 35 & 0  & 30 & 56 \\
                                                                                             & $n_{skip}=3$ & 56.10\% & 83.80\% & 93.00\% & 62.60\% & 53 & 6  & 41 & 6  & 27 & 49 \\
\bottomrule
\end{tabular}
\vspace{-2.0em}   % make gap between caption and
\end{table*}
\normalsize  % size of the font

In order to evaluate the actual tracking performance of the proposed framework using in-field surveillance videos, a custom tracking dataset is built using a 30-min long video (speed: 7FPS, resolution: 1280x720 px$^{2}$) collected at DRFC. The dataset (in MOT-16 format) includes 53 unique individuals partaking in activities over 9847 frames. This custom dataset can represent the sparsity/crowdedness and the moving patterns of the patrons at the study location. The proposed framework has satisfactory evaluation scores on the custom dataset and has a much lower ID switch number than other methods. The MOT metric \cite{bernardin2008evaluatingMOT} adopted to obtain the tracking performance evaluation scores (Table~\ref{tab:experiment_tracker}). In practice, tracking patrons in video can often generate more unique IDs due to false positives in the detection process. However these tracklets are usually short and are over a small number of frames. This study also tests a post processing step to filter out these tracklets. The threshold for filtering is set as 4, any tracklets that is shorter than 4 frames will be discarded. It is found this filtering step can reduce the total ID count to the ground truth making it a useful function in real practice for patron counting. The error in patron counting using both detection and tracking is about 7\% which is acceptable in an engineering application.

\subsection{Group Detection and Social Distance Measuring}

\begin{table}[bp]
\vspace{-1.2em}   % make gap between caption and
\centering
\caption{Group detection results on DRFC grouping dataset.}
\label{tab:experiment_grouping}
\footnotesize
\begin{tabular}{>{\centering\arraybackslash}p{1.1cm}>{\centering\arraybackslash}p{1.1cm}>{\centering\arraybackslash}p{0.9cm}>{\centering\arraybackslash}p{0.9cm}>{\centering\arraybackslash}p{0.9cm}>{\centering\arraybackslash}p{0.9cm}}
\toprule
Training dataset   &Testing dataset  & Window size    & Prcn  & Rcll  & F1 \\
\midrule
\textit{CBE}     & \textit{DRFC}         & 10s        & 83.72\%   & 82.33\%  & 83.02\%\\
\textit{CBE}     & \textit{stu003}       & 10s        & 91.91\%   & 81.27\%  &86.26\% \\
\bottomrule
\end{tabular}
\vspace{-2.0em}   % make gap between caption and
\end{table}
\normalsize

In order to validate the group detection algorithm in the field, a dataset is built by using a 1-hr video collected at DRFC. The dataset includes (manually annotated) unique IDs and trajectories for each individual in the real world coordinate system, and grouping information of the individuals. Instance-level bbox and segmentations on each frame are annotated firstly by detector and then manually revised. The custom grouping dataset consists of 154 unique individuals including cyclists, pedestrians, scooters, and skaters. There are 33 small groups consisting of 2 or 3 individuals and 86 singletons in total. Due to the lack of large grouping dataset with DRFC crowding scenario, the study uses the pre-trained results of the parameters (i.e. $\bm{\alpha}$ and $\bm{\beta}$ in Eq. \ref{eq:para_feature}) using a Structural SVM framework \cite{solera2015socially} on a public dataset for crowd analysis (\textit{Crowds-By-Examples (CBE)} dataset \cite{lerner2007crowds}). The group detection algorithm with the trained parameters is evaluated on DRFC grouping dataset. \textit{stu003} is one split of the \textit{CBE} dataset (with medium density crowd) including 406 unique individuals and 108 groups. For comparison, the evaluation results on \textit{stu003} split is also presented in Table~\ref{tab:experiment_grouping}. Although the parameters are not tuned for the DRFC environment, the group detection algorithm achieves comparable performance on the DRFC grouping dataset with an F1 score of 83.02\%. Training of the parameters will be carried out in the future once a large custom training dataset (at the study location) is available. A large custom training dataset is expected to get a more suitable combination of affinity features and can thus enhance the accuracy of the group detection algorithm in the field. 

\begin{figure} [tp]
\vspace{-1.0em}   % make gap between caption and
    \centering
    \begin{subfigure}[b]{0.23\textwidth}
    \centering
    \includegraphics[width=\textwidth]{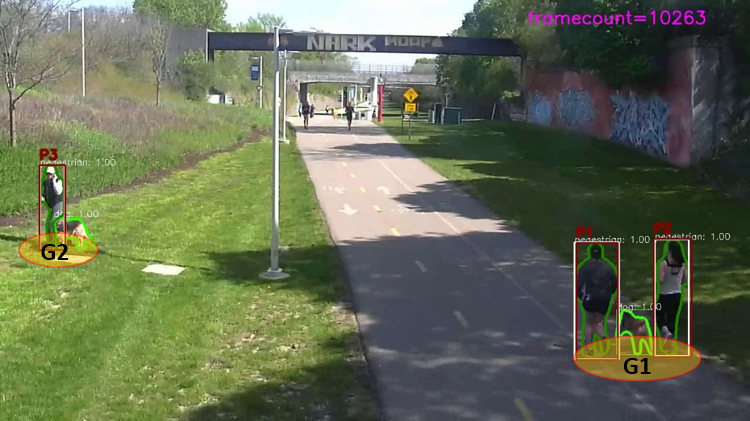}
    \vspace{-1.5em}   % make gap between caption and text smaller
    \caption{G1(1.2m),\\ G1-G2(12.7m)}    
    \label{fig:experiment_distance_a}
    \end{subfigure}
    \,
    \begin{subfigure}[b]{0.23\textwidth}  
    \centering 
    \includegraphics[width=\textwidth]{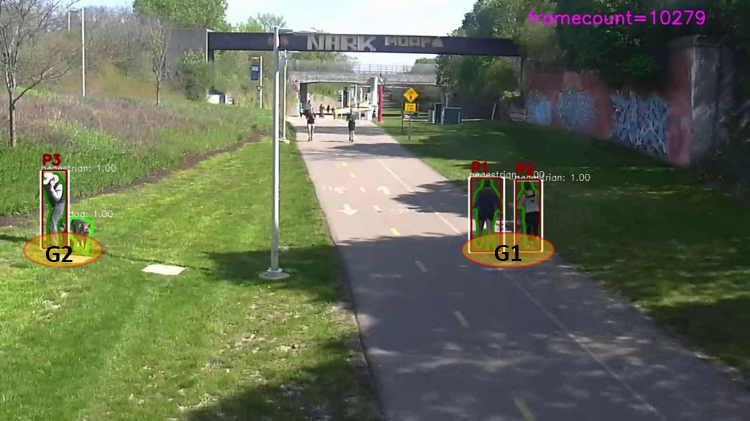}
    \vspace{-1.5em}   % make gap between caption and text smaller
    \caption{G1(1.0m),\\ G1-G2(9.8m)}    
    \label{fig:experiment_distance_b}
    \end{subfigure}
    \\
    \begin{subfigure}[b]{0.23\textwidth}
    \centering
    \includegraphics[width=\textwidth]{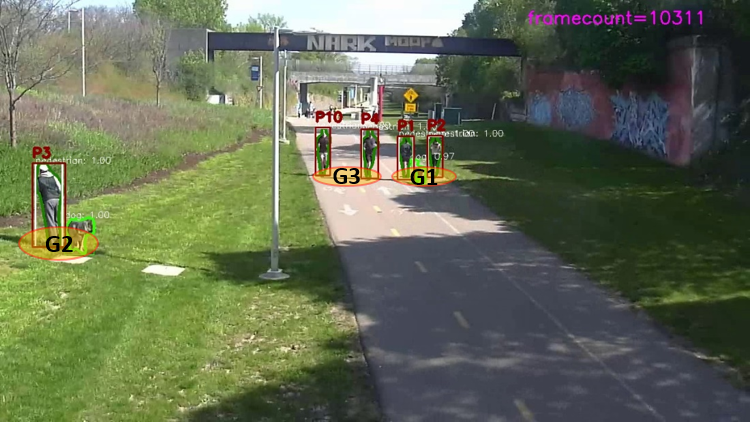}
    \vspace{-1.5em}   % make gap between caption and text smaller
    \caption{G1(1.2m),G3(1.9m),\\G1-G3(1.2m)}    
    \label{fig:experiment_distance_c}
    \end{subfigure}
    \,
    \begin{subfigure}[b]{0.23\textwidth}  
    \centering 
    \includegraphics[width=\textwidth]{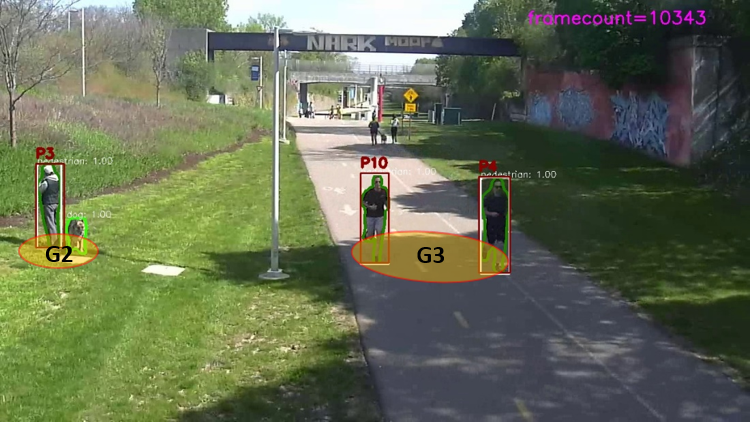}
    \vspace{-1.5em}   % make gap between caption and text smaller
    \caption{G3(2.5m),\\ G2-G3(7.4m)}    
    \label{fig:experiment_distance_d}
    \end{subfigure}
    \\
\caption{Examples of group clustering and social distance measuring on video frames on May 12, 2020.}
\vspace{-1.5em}   % make gap between caption and
\label{fig:experiment_distance}
\end{figure}

Fig.~\ref{fig:experiment_distance} shows examples of group clustering and measurements of social distances between different groups and inside each group. Pedestrians P1, P2, and P3 are firstly detected and tracked starting on frame-10263. P1 and P2 are walking upward while P3 is standing on the grass. Later, P4 and P10 enter the scene jogging downward. Based on the tracked trajectories, P1 and P2 are detected as group G1, P3 as group G2, and P4 and P10 as group G3. The diameter of groups and the distance between groups are monitored at different times/frames. For example, the diameter of group G1 is measured as 1.0-1.2m across different frames and the diameter of group G3 is measured as 1.9-2.5m across different frames. Although the diameter of G3 is a bit large, the similar trajectories and temporal relations between P4 and P10 ensure accurate group clustering. The closest distance between G1 and G3 occurs at frame-10311 (as show in Fig.~\ref{fig:experiment_distance_c}) when the distance is measured as 1.2m which is shorter than the suggested social distance of 2m. However, the total time for passing is very short and lasts only for a few seconds. Monitoring social distancing practices can be achieved automatically using the proposed sensing framework.

\subsection{Patron-Facility Contact Detection and Mask Detection}
Physical contact between patrons and park facilities can be inferred from the locations of detected patrons. A mapping module is developed to transform the detection and tracking results from pixel-wise coordinates to the real world coordinate system. Patrons are firstly detected and segmented by the detector on the video frame, then the bottom pixel coordinates ($u_{bot}$, $v_{bot}$) on the segmented contour corresponding to the bottom part of a patron (e.g. one foot for pedestrian or one tire for cyclist) are extracted (denoted as red dots in Fig.~\ref{fig:method_contact}). Based on the location information and other cues (e.g. aspect ratio of bbox, overlap between segmentations), physical contact events can be detected.

\begin{figure} [tp]
% \vspace{-1.0em}   % make gap between caption and
    \centering
    \begin{subfigure}[c]{0.467\textwidth}
    \centering
    \includegraphics[width=\textwidth]{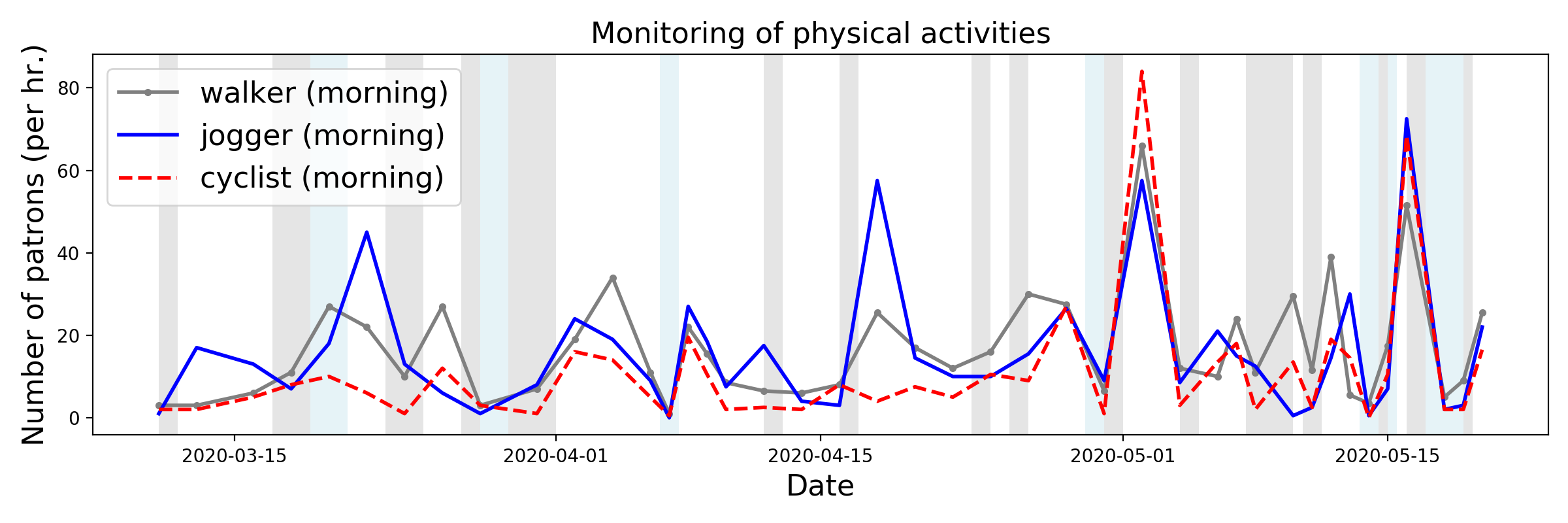}
    \vspace{-1.5em}   % make gap between caption and text smaller
    \caption{Patron activities}    
    \label{fig:experiment_count_patron}
    \end{subfigure}
    \\
    \begin{subfigure}[c]{0.49\textwidth}  
    \centering 
    \includegraphics[width=\textwidth]{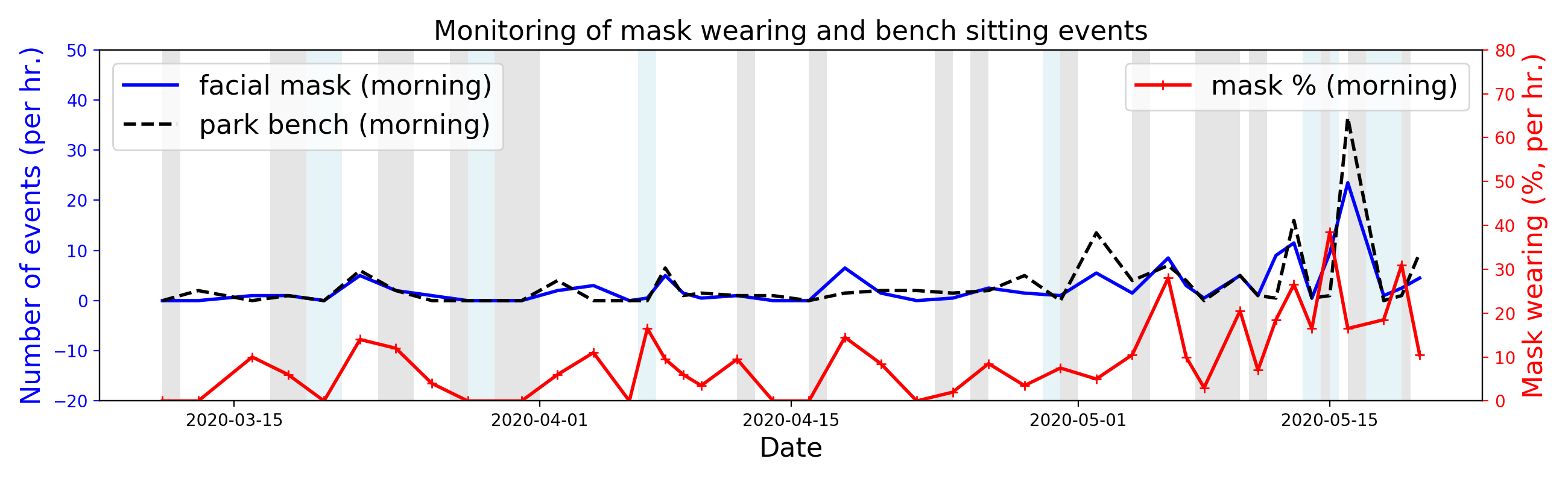}
    \vspace{-1.5em}   % make gap between caption and text smaller
    \caption{Contact and mask}    
    \label{fig:experiment_count_mask}
    \end{subfigure}
\\
\caption{Long-time monitoring of primary (a) patron activities and (b) hygienic practices at the Dequindre Cut of DRFC in mornings during COIV-19 pandemic (from Mar. 11 to May 19 2020). Note: the white background represents sunny, gray shade represents overcast, and light blue shade represents raining.}
\vspace{-1.5em}   % make gap between caption and
\label{fig:experiment_couting}
\end{figure}

In order to train and validate the proposed CNN-based mask detector/classifier, a custom dataset is built by collecting cropped images of patrons wearing and not wearing facial masks at DRFC during COVID-19 pandemic. The images are manually curated from surveillance camera footage and consist of 1,580 positive examples (cropped faces with facial coverings) and 3,720 negative examples (cropped faces without facial coverings and arbitrary crops from the surveillance footage). Of the 5,300 image test-set, 4,500 images are used for training while 800 images are withheld for validation. For training, a batch size of 325 is adopted with 45 epochs. Adaptive moment estimation (Adam) \cite{kingma2014adam} and a learning rate of 0.07 are used for training. The evaluation  yields a precision of 94\%, a recall of 80\%, and an F1 score of 86\% on the testing dataset. The performance is satisfactory considering the small size of the head portion of patrons and the low-resolution characteristics of the surveillance images.

The proposed proof-of-concept study also reports some applawications in automatic monitoring of patron activities and hygienic practices among patrons in a two-month period (Mar 11 - May 19, 2020). Video data from one surveillance camera at the Dequindre Cut is used to study the primary physical activities in the mornings from 10:00 to 12:00. It is found that most of the spikes in Fig.~\ref{fig:experiment_count_patron} occur during sunny days while low activity level is usually associated with gloomy or rainy weather conditions. The result shows the overall physical activities at DRFC (e.g. walking, jogging, and bicycling) have a strong correlation with weather conditions. Although fluctuating during the time, the overall trend of wearing facial masks in POS or at least at DRFC can be revealed from Fig.~\ref{fig:experiment_count_mask}. The conclusion is drawn by observing the increasing of both total number and rate of patrons who wear facial masks (the highest rate is about 40\% around May 15). Regarding the use of park benches, it seems that patrons at DRFC have not been avoiding them and are not following recommendations to avoid physical contact with hard surfaces in public spaces.

\section{Conclusion} \label{sec:conclustion}
This paper presents a proof-of-concept study of an automatic sensing framework to measure social distancing and hygienic practices in public open spaces during COVID-19 pandemic. The multi-task framework integrates multiple function modules to achieve patron sensing and activity recognition, social distance monitoring between social groups, and facial mask detection. The function modules for various tasks have been validated using multiple custom datasets with manual annotations on DRFC videos. The proposed sensing framework is at service at DRFC and some preliminary results from in-field applications are also demonstrated. More studies on improving the framework and more applications in patron activity sensing during the COVID-19 will be carried out in future. The authors envision wider applications of the method in other public open spaces and different built environments.  

\section*{Acknowledgement}
The support from the National Science Foundation (NSF) under grant \#1831347 is gratefully acknowledged.

{\small
\bibliographystyle{ieee_fullname}
\bibliography{social_distance}
}

\end{document}

%%%%%%%%% TITLE
\title{Supplemental Material: An Autonomous Approach to Measure Social Distances and Hygienic Practices during COVID-19 Pandemic in Public Open Spaces}

\author{Peng Sun\\
University of Central Florida\\
{\tt\small peng.sun@ucf.edu}

\and
Gabriel Draughon\\
University of Michigan\\
{\tt\small draughon@umich.edu}

\and
Jerome Lynch\\
University of Michigan\\
{\tt\small jerlynch@umich.edu}
}

\maketitle

\newpage
\begin{appendices}
% \appendix
\setcounter{figure}{0}
\setcounter{equation}{0}
\counterwithin{figure}{section}
\counterwithin{equation}{section}

\section{Detector} \label{append:detector}
\begin{figure}[bp]
% \vspace{-1.5em}   % make gap between caption and
\centering
\includegraphics[width=1.0\linewidth]{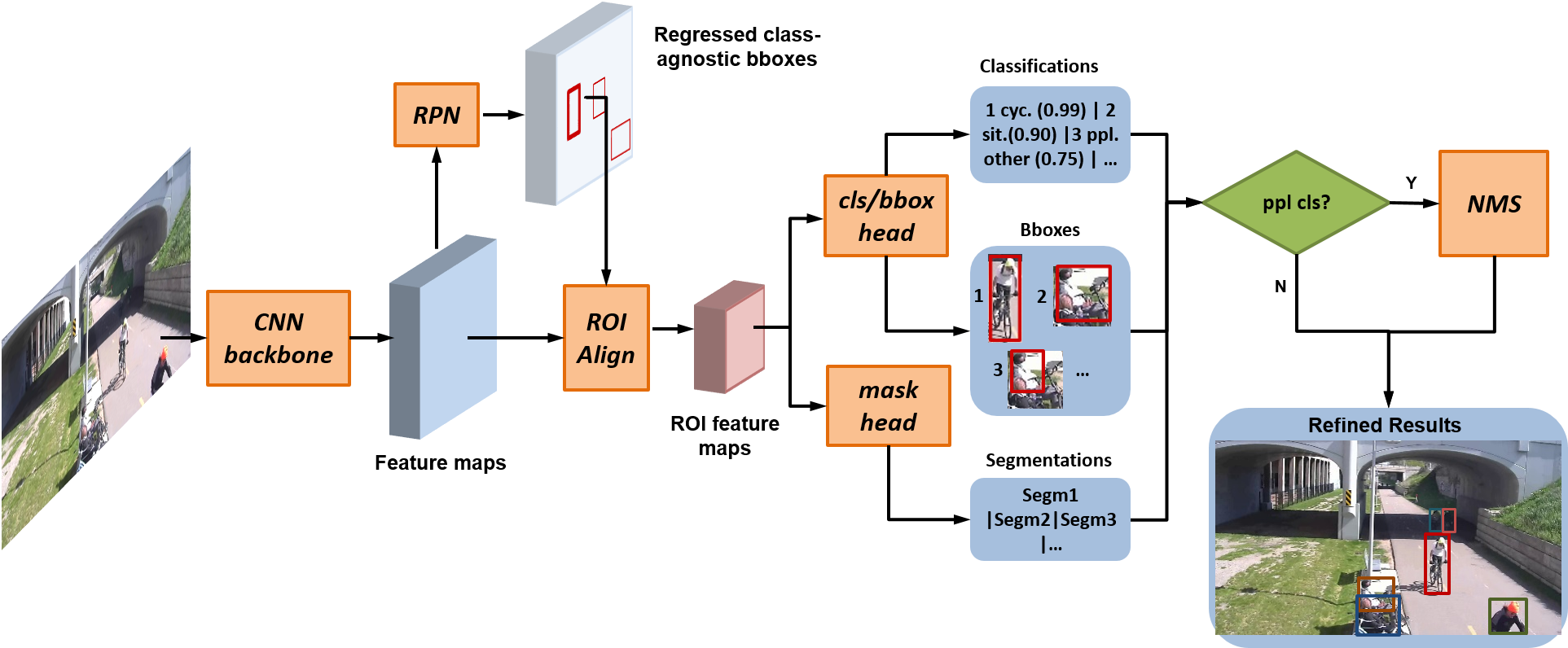}
\caption{Schematic of the Mask R-CNN detection model with CNN + FPN backbone. Note: adding the non-maximum suppression (NMS) for human classes can remove the double detection of patrons in POS.}
% \vspace{-1.0em}   % make gap between caption and
\label{fig:method_detector}
\end{figure}

In this study, the RPN is trained together with the other parts of the Mask R-CNN model  (Fig.~\ref{fig:method_detector}) in an end-to-end fashion. Including the RPN loss ($L_{rpn}$), the total loss function for the end-to-end training process for Mask R-CNN is defined as:
\begin{equation}
    L_{maskrcnn}= (L_{rpn\_obj}+ L_{rpn\_bbox}) +(L_{cls}+L_{bbox}+L_{mask})
\end{equation}

\noindent where $L_{rpn\_obj}$ and $L_{rpn\_bbox}$ represent the loss for the classifier (i.e. object or not object of interest) and the loss for the bbox regression within the RPN, respectively; while $L_{cls}$, $L_{bbox}$, and $L_{mask}$ represent the losses for the classifier, bbox regressor, and instance segmentation within the detection heads, respectively. 

For comparison, the study will also show the performance of some one-stage detection methods (e.g. RetinaNet \cite{lin2017RetinaNet} or YOLOV3 \cite{redmon2018yolov3}). The loss functions of the two methods are similar and can be generally defined as:
\begin{equation}
    L_{single}= L_{obj} + L_{cls} + L_{bbox}
\end{equation}
\noindent where $L_{obj}$, $L_{cls}$, and $L_{bbox}$, represent the losses for objectiveness, classifier, and bbox regressor, respectively. Note: details in the loss functions are different between RetinaNet and YOLOV3 (e.g. RetinaNet uses focal loss).

\section{Tracker} \label{append:tracker}
The Kalman filter recursively generates the current state ($X_{k}$) and the process covariance matrix ($P_{k}$) on the $k$-th video frame by using the previous state ($X_{k-1}$), the previous covariance matrix ($P_{k-1}$), and the newly arrived measurement ($Y_{k}$). In this study, the state $X=\{x_{c}, y_{c}, a, h, \dot{x_{c}},\dot{y_{c}},\dot{a},\dot{h}\}^{T}$ is defined to include the center location, aspect ratio, height and their velocities of the tracked bboxes. 

The projection of the state into measurement is:
\begin{equation}
    Y_{k}=C X_{km} + Z_{m}
\end{equation}
\noindent where $Y_{k}$ is the observation of the measured state ($X_{km}$),  $Z_{m}$ is the measurement uncertainty, and $C$ is the observation matrix from $X_{km}$.

The updated prediction (i.e. $X_{k}$ and $P_{k}$) from the Kalman filter is compared with the detection results on $k$-th video frame. The squared Mahalanobis distance between the predicted Kalman states of one track (e.g. $i$-th track) and one newly detection (e.g. $j$-th detection) on the current frame is computed as:

\begin{equation}
    d_{mot}(i,j)=(det_{j}-Y_{i})^{T} P_{i}^{-1}(det_{j}-Y_{i})
\label{eq:d_mot}
\end{equation}

\noindent where $det_{j}$ is the $j$-th bbox detection (e.g. a vector of the bbox locations, aspect ratio, height, and the corresponding changing rates), $(Y_{i}, P_{i})$ denotes the projection of the $i$-th track distribution into measurement space. $Y_{i}$ is the predicted observation and $P_{i}$ is the covariance matrix.

In addition to the motion related information, appearance information is utilized for the comparison between a detection and a track in order to reduce the occurrence of ID switches which happens due to a short time of of occlusion. Descriptors of the detected objects (i.e. cropped images by the detected bbox) are generated using the CNN. The feature vector or descriptor has a standard length of 128. The appearance metric measures the smallest cosine distance between the $i$-th track and $j$-th bbox detection and it is expressed as:

\begin{equation}
    d_{app}(i,j)=min \{ 1- \bm{r}_{j}^{T}\bm{r}_{l}^{(i)} \; | \; \bm{r}_{l}^{(i)} \in  \CMcal{R}_{i} \}
\label{eq:d_app}
\end{equation}

However, the influence of the motion information is incorporated when dealing with the gating parameter. Because when the $d_{mot}(i,j)$ is too large, the corresponding $d_{comb}(i,j)$ is set as infinity (i.e. $10^{5}$ for computation) to filter out the unqualified matches $(i,j)$. A strict requirement for deciding the admissibility between a track and a detection: both the requirement for motion and the appearance should be satisfied to guarantee the final admissibility.

The MOT metric for evaluation includes: multiple object tracking accuracy (MOTA), multiple object tracking precision (MOTP), mostly tracked targets (MT), mostly lost targets (ML), and the ID switch number (IDs) etc. MOTA allows for objective comparison by considering the overall performance of a tracker (and a detector if custom detection is used), including misses, false positives, and mismatches over all frames.

\begin{equation}
  \text{MOTA} = 1 - \frac{\sum_{t}(FN_{t}+FP_{t}+IDs_{t})}{\sum_{t}GT_{t}}
\end{equation}
\noindent where $t$ is the frame index, $g_{t}$ is the number of ground truth objects at $t$. $FN_{t}$, $FP_{t}$, and $IDs_{t}$ are the numbers of misses, false positives, and mismatches at $t$, respectively.

MOTP in this study follows the MOT Challenge benchmarks definition rather than original definition in \cite{bernardin2008evaluatingMOT}. MOTP is a measure of localization precision and in practice it mostly quantifies the localization performance of the detector not the tracker.

\begin{equation}
  \text{MOTP} = 1 - \frac{\sum_{t,i} d_{t,i}}{\sum_{t}c_{t}}
\end{equation}
\noindent where $c_{t}$ denotes the number of matches, and $d_{t,i}$ is the bbox overlap of target $i$ with its assigned ground truth object. 

\section{Mapping Module} \label{append:mapping}
The intrinsic matrix ($A$) can be computed using a chessboard at various positions. The joint rotation-translation matrix $[\bm{R|t}]$ is the matrix of the extrinsic parameters.  The relationship between a location in the WCS ($\{X, Y, Z\}$) and the location in the CCS ($\{X_c, Y_c, Z_c\}$) is expressed in the equation below:

\begin{equation}
	Z_c \begin{bmatrix} X_c' \\ Y_c' \\ Z_c' \end{bmatrix} =	
	\begin{bmatrix} X_c \\ Y_c \\ 1 \end{bmatrix} = 
	\bm{R} \begin{bmatrix} X \\ Y \\ Z \end{bmatrix}  + \bm{t}
\end{equation}

\noindent where $X_c'=X_c/Z_c$ and $Y_c'=Y_c/Z_c$.

Theoretically, the relationship between a point in the CCS and the same point in the PCS is straightforward and can be converted between the two coordinate systems using the $\bm{R}$ rotation matrix and $\bm{t}$ translation vector. However, in practice, camera lenses typically have radial and tangential distortion\cite{ma2012imagemodel} which can be modeled and corrected with radial and tangential distortion coefficients. The intrinsic matrix of a camera can be obtained by taking multiple images (e.g., 100-200) of a chessboard in various positions and orientations. Because the intrinsic parameters of a camera do not change, the computed intrinsic matrix $\bm{A}$ can be stored and re-used. However, the external parameters (e.g. joint rotation-translation matrix $\bm{[R|t]}$) have to be calculated for each new scene. These calculations are used to optimize the location and pose of the camera to ensure object locations (e.g., the patterns of marks printed on a road) in the 2D PCS and 3D WCS are well matched. Assuming a flat road plane with $Z_{bot}=0$, the world coordinates of the bottom pixel ($X_{bot}$, $Y_{bot}$) can be solved using the camera's intrinsic matrix $\bm{A}$ and rotation-translation matrix $\bm{[R|t]}$. 

\section{Crowd Analysis in Literature} 
\label{append:group_anaysis}
Analysis on crowd behavior has been researched in the past few decades for many applications, such as, to study people's physical activity in urban green spaces \cite{hunter2015impact}, to  enhance video surveillance \cite{sreenu2019intelligent}, and to measure crowd statistics (e.g. counting and density) \cite{oghaz2019content} from static images. In the past decade, a number of studies have utilized deep learning methods for crowd analysis \cite{zhan2008crowd} including, crowd counting and density estimation, crowd motion detection, crowd tracking, and crowd behaviour understanding. Instead of analyzing large crowd dynamics in densely packed scenes (e.g., count$>$100 or even count$>$1000) this study will mainly focus on space usage patterns across varied micro-environments within POS. Overly crowded scenes are rare occurrences in typical U.S. community parks, therefore large crowd dynamics are beyond our scope of study. Rather this study focuses on the distances, counting, and spatio-temporal distribution of entities within and between small groups (e.g. count$<$100). 

\section{Group Detection and Features} \label{append:group_detection}
The aggregated pairwise feature adopted in the study \cite{ge2012vision} can be calculated as follows:
\begin{equation}
\begin{split}
    f_{fram}^{t}(i,j) = & \lambda \CMcal{N}_{min-max} (\norm{ s_{i}^{t}-s_{j}^{t} }^2) \\ 
    & + (1-\lambda) \CMcal{N}_{min-max} (\norm{ v_{i}^{t}-v_{j}^{t} }^2)
\end{split}
\label{eq:feature_ge}
\end{equation}

\noindent where $\CMcal{N}_{min-max}(\cdot)$ is a min-max normalization operator for velocity and distance within a range of [0,1], $\lambda \in [0,1]$ is a parameter to tune the proportions of location in the aggregated feature, and qualified pairs of $(i,j)$ are filtered in using the pre-defined thresholds of distance and velocity differences (i.e. $\tau_s$ and $\tau_v$).

This designed feature $f_{fram}^{t}(i,j)$ in Eq. \ref{eq:feature_ge} is computed from static proximity and the velocity (both direction and magnitude) at each time instance, or per video frame. 

The proxemics feature can be treated with a Gaussian Mixture Model (GMM) using spatial quantizations based on Hall's boundaries \cite{hall1966hidden} for pedestrians. The boundaries between people were categorized as intimate (0-0.5 m), personal (0.5-1.2 m), social (1.2-3.7 m), and public boundary (3.7-7.6 m). The public boundary metric (7.6m) is not considered in this study since it is too large accurately discern in person or over video footage.
Hence, the study chooses the first three boundaries for building the GMM:
\begin{equation}
\begin{aligned} 
    GMM(s_{i}^{t}, s_{j}^{t}) &= \frac{1}{3}\sum_{m}^{3} \CMcal{N}(\bm{s}_{i}^{t}-\bm{s}_{j}^{t}|\bm{0},\bm{\sigma}_m) \\
                            &= \frac{1}{3}\sum_{m}^{3} \CMcal{N}(x_{i}^{t}-x_{j}^{t}|0,\sigma_m) \cdot \CMcal{N}(y_{i}^{t}-y_{j}^{t}|0,\sigma_m)  \\
\end{aligned} 
\end{equation}
\noindent where $\CMcal{N}(\cdot|\mu,\sigma)$ is a normal distribution and $\sigma_m$ is the $m$-th standard deviation set of the boundaries (i.e. $\sigma_1$ =0.5m, $\sigma_2$ =1.2m, $\sigma_3$ =3.7m).

In order to compute the average proxemics feature between two trajectories, multiple frames should be considered:
\begin{equation}
   f_{1}^{k}(i,j) = \frac{1}{N_f} \sum_{t \in T_{k}} GMM(\bm{s}_{i}^{t}, \bm{s}_{j}^{t})
\end{equation}
\noindent where $N_f$ is the number of the frames when $i$ and $j$ both exist (tracked in tracking module) on a frame with the condition that $t \in \CMcal{T}_{k}$. Averaging the value across frames improves the robustness.

The trajectory shape similarity feature compares the trajectories (e.g. $\CMcal{T}_{i}$ and $\CMcal{T}_{j}$) of a pair of time invariant series.The Dynamic Time Warping (DTW) algorithm \cite{berndt1994dtw} is used to map one trajectory $T_{i}$ to another trajectory $T_{j}$, driving distance between the two to a minimum. DTW allows two trajectories that are similar in shape but locally out of phase to align.

The most aligned patterns are found by minimizing the cumulative cost $\gamma_{ij}$ recursively of any path through the distance matrix $D_{ij}$:
\begin{equation}
\begin{split}
  \gamma_{ij}^{k}(m,n) = &(D_{ij})_{mn} + \minimize_{1<m<M,1<n<N } [\gamma _{ij}^{k}(m-1,n),\\
  &\gamma _{ij}^{k}(m-1,n-1),\gamma _{ij}^{k}(m,n-1) ]
\end{split}
\end{equation}
\noindent where the element $(D_{ij})_{mn}$ within the distance matrix $D_{ij} \in \mathbb{R}^{M \times N}$, is used to encode the squared Euclidean distance of the $m$-th element of trajectory $T_{i}$ (with length of $M$) and the $n$-th element of trajectory $T_{j}$ (with length of $N$). Note the recursive process starts from $\gamma _{ij}^{k}(1,1)=(D_{ij})_{11}$.

Once DTW is performed, the trajectory shape similarity feature is constructed as:
\begin{equation}
   f_{2}^{k}(i,j) = \gamma_{ij}(M,N)/\text{max}(M,N), t \in \CMcal{T}_{k}.
\end{equation}

The third feature, motion causality, is extracted using the econometric model of Granger causality \cite{granger1969investigating}. The fourth feature, path convergence, is extracted using a heat map-based method \cite{lin2013heat} to holistically model groups. For details to extract the motion causality and path convergence, please refer to \cite{solera2015socially}.

\section{Data Privacy Protection}
The automation of sensing framework is designed to protect data privacy. First, the project with Detroit riverfront parks has been performed under an 
Institutional Review Boards (IRB) human-subject review process. In addition, the cameras in current use at the Detroit riverfront lack sufficient resolution for facial recognition. Second, the parks advertise the presence of the surveillance cameras to ensure that the patrons are fully aware of their presence. Furthermore, the sensing framework is designed to only save an anonymized version of the patrons with raw images not saved once the system is fully trained. As an added layer of precaution, the associated detection and tracking datasets have removed facial features from the images to ensure public privacy. The algorithmic framework would extract park users detection and activity results as a completely anonymized representation of park users to
long-term storage. 
\end{appendices}

{\small
\bibliographystyle{ieee_fullname}
\bibliography{social_distance}
}